\documentclass[letterpaper, 10 pt, journal, twoside]{IEEEtran}

\IEEEoverridecommandlockouts

\usepackage[T1]{fontenc}
\usepackage[utf8]{inputenc}
\usepackage[pdftex]{graphicx}
\graphicspath{{./images/}, {./imgs/}, {./imgs_updated/}}
\DeclareGraphicsExtensions{.png, .pdf}

\usepackage{amsmath} %
\usepackage{amssymb}  %
\usepackage{url}
\usepackage{xcolor}
\usepackage{xspace}
\usepackage{enumitem}
\usepackage{multirow}
\usepackage{array}
\usepackage{tabularx}
\usepackage{textcomp}
\usepackage{mathtools}
\usepackage{bm}
\usepackage{dsfont}
\usepackage{cite}
\usepackage{microtype}

\usepackage{eso-pic}

\usepackage{footnote}
\makesavenoteenv{tabular}
\makesavenoteenv{table}

\usepackage[capitalise]{cleveref}

\DeclareMathOperator*{\argmax}{arg\,max}

\makeatletter
\let\MYcaption\@makecaption
\makeatother
\usepackage[font=footnotesize]{subcaption}
\makeatletter
\let\@makecaption\MYcaption
\makeatother

\begin{document}

\title{Spiking Neural Networks for Visual Place Recognition via Weighted Neuronal Assignments} %

\author{Somayeh Hussaini, \textit{Student Member, IEEE}, Michael Milford \textit{Senior Member, IEEE},\\and Tobias Fischer, \textit{Member, IEEE}\vspace{-.5cm}
\thanks{Manuscript received: September, 9, 2021; Revised December, 1, 2021; Accepted January, 29, 2022.}
\thanks{This paper was recommended for publication by Editor T. Ogata upon evaluation of the Associate Editor and Reviewers' comments.
This work was supported by the Australian Government via grant AUSMURIB000001 associated with ONR MURI grant N00014-19-1-2571, Intel Research via grant RV3.248.Fischer, and the Queensland University of Technology (QUT) through the Centre for Robotics.}
\thanks{The authors are with the QUT Centre for Robotics, Queensland University of Technology, Brisbane, QLD 4000, Australia {\tt\footnotesize somayeh.hussaini@hdr.qut.edu.au}} 
\thanks{Digital Object Identifier (DOI): 10.1109/LRA.2022.3149030}
}

\markboth{IEEE Robotics and Automation Letters. Preprint Version. Accepted January, 2022}{Hussaini \MakeLowercase{\textit{et al.}}: Spiking Networks for Place Recognition via Weighted Neuronal Assignments}  %

\AddToShipoutPicture*{%

\AtTextUpperLeft{%

\put(-3.5,10){

\begin{minipage}{\textwidth}

\scriptsize

\MakeUppercase{Preprint version; final version available at} \url{http://doi.org/10.1109/LRA.2022.3149030}

\end{minipage}}%

}%

}

\maketitle

\begin{abstract}
Spiking neural networks (SNNs) offer both compelling potential advantages, including energy efficiency and low latencies and challenges including the non-differentiable nature of event spikes. Much of the initial research in this area has converted deep neural networks to equivalent SNNs, but this conversion approach potentially negates some of the advantages of SNN-based approaches developed from scratch. One promising area for high-performance SNNs is template matching and image recognition. This research introduces the first high-performance SNN for the Visual Place Recognition (VPR) task: given a query image, the SNN has to find the closest match out of a list of reference images. At the core of this new system is a novel assignment scheme that implements a form of ambiguity-informed salience, by up-weighting single-place-encoding neurons and down-weighting ``ambiguous'' neurons that respond to multiple different reference places. In a range of experiments on the challenging Nordland, Oxford RobotCar, SPEDTest, Synthia, and St Lucia datasets, we show that our SNN achieves comparable VPR performance to state-of-the-art and classical techniques, and degrades gracefully in performance with an increasing number of reference places. Our results provide a significant milestone towards SNNs that can provide robust, energy-efficient, and low latency robot localization.%
\end{abstract}

\begin{IEEEkeywords}
Neurorobotics, Localization, Autonomous Vehicle Navigation
\end{IEEEkeywords}

\section{Introduction}
\IEEEPARstart{S}{piking} neural networks (SNNs) differ in several aspects from conventional artificial neural networks (ANNs)~\cite{maass1997networks}. A neuron's output within an ANN is a real number computed by a non-linear function of the sum of the inputs. This computation is different from the spikes (also called events) in SNNs, which carry a single bit of information~\cite{maass1997networks,davies2021advancing,maass2015spike}. Another difference is the learning algorithm: ANNs rely on the backpropagation of error signals, which in the standard implementation does not have an analogue in the brain~\cite{maass2015spike,WHITTINGTON2019235}. In contrast, the brain uses unsupervised learning methods like Spike Time Dependent Plasticity (STDP), which are also used to train SNNs~\cite{markram2012spike}. Although biological insights or fidelity are not the primary focus of this paper, the premise is that SNNs are better suited to study the operation of biological systems and for (in-)validating hypotheses~\cite{maass1997networks,maass2015spike}.

\begin{figure}[t]
\centerline{\includegraphics[width=0.97\columnwidth]{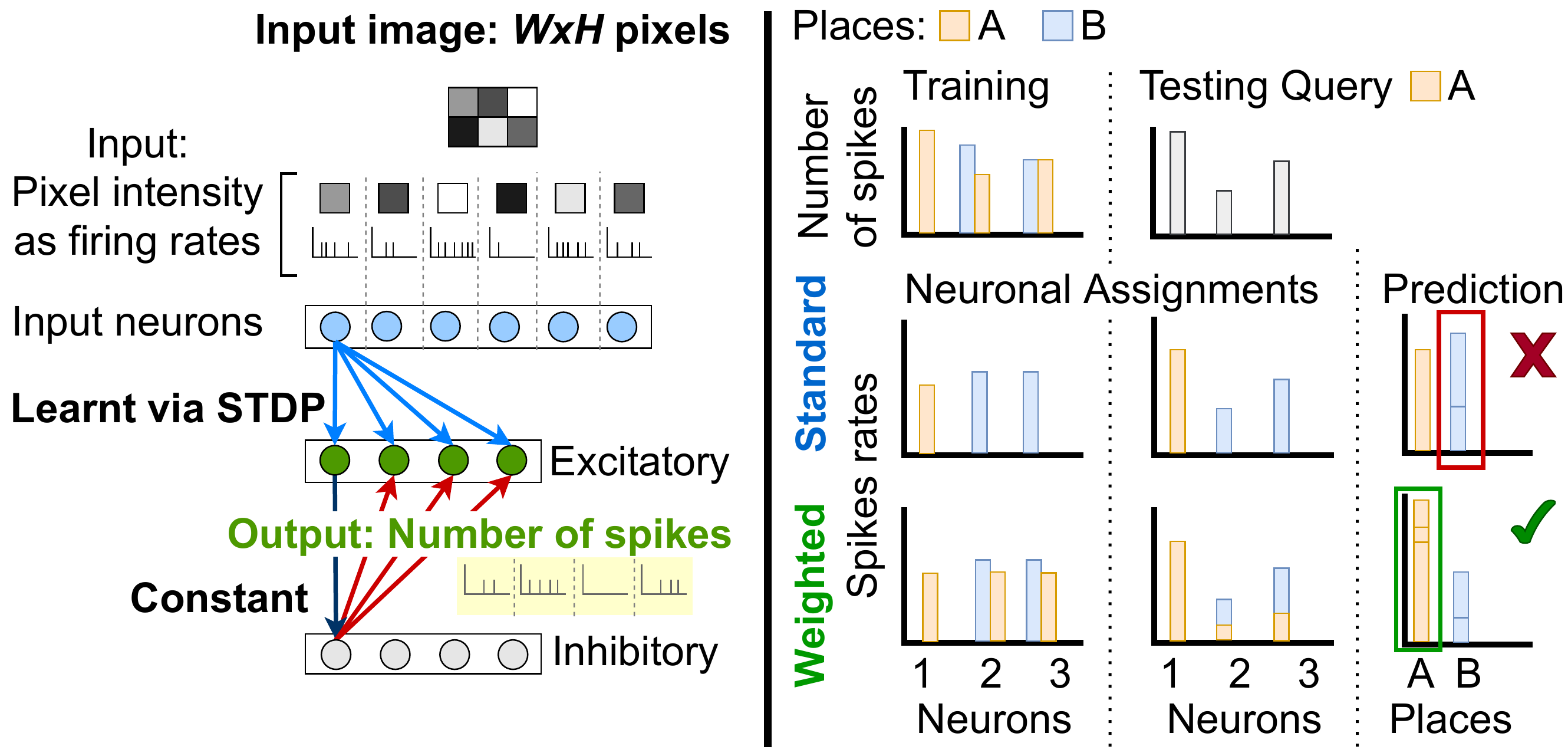}}
\vspace*{-0.15cm}
\caption{A Spiking Neural Network (SNN) model for Visual Place Recognition (VPR): Left: SNN model. The arrows in each layer depict the connection of one neuron to the neurons of the next layer. For simplicity, we only show the connections of one neuron in each layer. Right: Neuronal responses in training and testing. The standard assignments assign neurons to place labels based on the highest responses, leading to a wrong prediction. The weighted assignments consider the ambiguity of neuronal responses by scaling neurons that have learned multiple places, leading to a correct prediction.  \vspace*{-0.3cm}}
\label{fig:motivation}
\end{figure}

Recently, SNNs have gained popularity in the robotics community~\cite{tieck2017towards,tang2018gridbot,tieck2018controlling,tang2019spiking,kreiser2018pose,kreiser2020chip,lele2021end,vitale2021event}, due to potential advantages over ANNs including low power consumption and low latencies. The community has access to a growing neuromorphic hardware set, such as Intel's Loihi~\cite{davies2021advancing}. SNNs have e.g.~been used in robot grasping~\cite{tieck2017towards}, robot navigation~\cite{tang2018gridbot}, and motion planning~\cite{tieck2018controlling}. They are however not yet widely competitive; the resulting accuracy is generally lower compared to conventional non-spiking approaches. Current practical use (beyond scientific curiosity) is limited to narrow cases like high-speed or low power scenarios, or where closed-loop control with low latencies is required \cite{davies2021advancing}. 

In this paper, we take a significant step towards bridging the performance divide between conventional and SNN-based techniques for the Visual Place Recognition task (VPR). VPR is critical in enabling long-term mobile robot autonomy in challenging environments: it can aid in the global re-localization of mobile robots, minimize model errors by acting as a closed-loop detection algorithm, and can be used as a component of robust localization and mapping algorithms~\cite{Lowry2015,Garg2021}. 
\vspace{0.1cm} %

\noindent Our research here makes the following contributions:
\begin{enumerate}
    \item We introduce the first SNN for the place recognition task, combining leaky integrate and fire neurons within a lightweight, biologically plausible SNN to represent place templates using unsupervised learning via STDP.
    \item We introduce multiple novel neuronal assignment schemes that decode all neurons' spike responses and weight them based on the specificity or ambiguity of a single neuron's response to learning places.
    \item We demonstrate competitive performance in terms of recall at 100\% precision when comparing our SNN to conventional VPR baselines, namely NetVLAD~\cite{Arandjelovic2018} and the Sum-of-Absolute Differences (SAD)~\cite{milford2012seqslam}, on the Nordland~\cite{sunderhauf2013we}, Oxford RobotCar~\cite{RobotCar}, SPEDTest \cite{chen2018SPED}, Synthia \cite{ros2016synthia}, and St Lucia \cite{Glover2010StLucia} datasets.
    \item We present ablation studies showing graceful performance degradation for a fixed network size as the number of places to be learned increases. %
\end{enumerate}

\newcommand{\code}{\texttt}

To foster future research, we make the code to replicate our experiments available with additional supporting materials and ablation studies: \texttt{https://github.com/QVPR/VPRSNN}.

\section{Related Works}
\label{sec:relatedworks}

Here we review recent Spiking Neural Network research with an emphasis on robotic-related applications in~\Cref{subsec:RelSNN}, and an overview of Visual Place Recognition in~\Cref{subsec:RelVPR}.

\subsection{Spiking Neural Networks}
\label{subsec:RelSNN}

The field of neuromorphic computing concerns the development and advancement of neuromorphic hardware, sensors, and associated algorithms. Diehl and Cook~\cite{diehl2015unsupervised}'s two-layer Spiking Neural Network (SNN) model demonstrated the potential capability of SNNs in solving pattern recognition tasks such as digit recognition. More broadly, SNNs have been evaluated in various robotics applications~\cite{tieck2017towards,tang2018gridbot,tieck2018controlling,tang2019spiking,kreiser2018pose,kreiser2020chip,lele2021end,vitale2021event,zhu2020spatio}, including head pose tracking and scene understanding \cite{kreiser2020chip}, prey tracking \cite{lele2021end}, and control of high-speed unmanned aerial vehicles \cite{vitale2021event}.

Biologically-inspired SNNs have in particular been developed for robot navigation-related tasks including Simultaneous Localization and Mapping (SLAM)~\cite{cadena2016past}. These works include RatSLAM, originally proposed by \cite{milford2004ratslam}, on a neuromorphic hardware using place, grid and border cells \cite{galluppi2012live}, place cells for 3-dimensional path planning \cite{steffen2020networks}, spatio-temporal memory~\cite{zhu2020spatio}, and pose estimation and mapping \cite{kreiser2018pose} demonstrated on simulated environments. Similar works consider mapping constraint indoor environments which include an energy-efficient uni-dimensional SLAM \cite{tang2019spiking}, and a neuromorphic-based navigation method on an autonomous robot \cite{tang2018gridbot}. Our work is thus complementary to these works on SLAM by providing a loop closure detection algorithm in long-term and large-scale mapping of real-world environments.

Recently, event cameras have gained popularity as a sensing modality to use for SLAM. Event cameras output pixel-level intensity changes, which can result in significant information reduction compared to conventional images \cite{Gallego2019}.
Event-based input has been used for robotic navigation \cite{milford2015place, weikersdorfer2013simultaneous}, and in combination with other sensors for a quadrotor navigation system \cite{vidal2018ultimate}. Other works for robot navigation have used event-based data as input to SNNs. These works tackled estimation of a robot's orientation for path integration \cite{kreiser2018neuromorphic}, and map formation and pose estimation tasks \cite{kreiser2020error} which are deployed on neuromorphic devices and are applicable to SLAM solutions.

While SNNs and related neuromorphic hardware have promising properties, there is often still a performance gap when compared to modern conventional learning- and algorithmic-based techniques. Due to the non-differentiable neuronal dynamics of spiking neurons, supervised techniques such as back-propagation cannot be directly applied. Recent works \cite{renner2021backpropagation, lee2020enabling} have proposed algorithms that approximate backpropagation, and demonstrated use on Intel's Loihi. 

Another common approach to implement high-performing SNNs is through converting Artificial Neural Networks (ANNs) to their equivalent SNNs~\cite{rueckauer2017conversion,stockl2021optimized}. These models can achieve comparable performance to their ANN predecessors, but may not necessarily be an ideal form of SNNs in the long term due to not fully exploiting the advantages of computational SNNs; in particular, unsupervised methods and methods that approximate backpropagation can lead to spike-timing codes with more optimized latency and energy efficiency~\cite{davies2021advancing}.

\subsection{Visual Place Recognition} 
\label{subsec:RelVPR}

Visual place recognition (VPR) is the task of recognizing whether a place has been already visited given a current image of the place, and prior knowledge of previously visited places \cite{Lowry2015,Garg2021}. The VPR problem can be considered as a template matching problem, where the closest match of a query image is selected from a reference set of places that can be captured from significantly different viewpoints and appearances.

Early works for the VPR problem implemented hand-crafted feature representations -- either from local points or by aggregating features over the whole image~\cite{cummins2008fab,jegou2010vlad,milford2012seqslam}. Many of those hand-crafted approaches have subsequently been turned into deep neural networks, enabling feature representation learning.

The most popular of these approaches is arguably NetVLAD \cite{Arandjelovic2018}, based on the Vector of Locally Aggregated Descriptors (VLAD)~\cite{jegou2010vlad}, which implements a pooling layer with differentiable operations to allow end-to-end training. NetVLAD has formed the foundation for a substantial number of follow-on systems \cite{hausler2021patch, yu2019spatial}, but is still competitive performance-wise, and is used as a baseline in our experiments (\Cref{ER_SOTA}).

One conceptually relevant method is Bayesian Selective Fusion~\cite{molloy2020intelligent}, which finds the best match using \emph{several sets of reference images}, similar to the use of multiple reference sets here to train the SNN. Other relevant work includes that of Frady et al.~\cite{frady2020neuromorphic} who leverage the advantages of Intel's neuromorphic Loihi processor~\cite{davies2021advancing} with a scalable k-Nearest Neighbor (kNN) search algorithm. This work is of particular importance, as it shows that SNNs, solving similar tasks to ours, can achieve similar accuracy as CPU implementations, while having low latency ($\approx$3ms) \emph{and} high throughput (>300 queries/s) \emph{and} low power ($\approx$10W).

\section{Methodology: Preliminaries}
\label{sec:methods}

Implementation of SNN-based models can pose a significant technical learning curve, especially for researchers coming from a conventional neural network background. In the following sections, we attempt to provide sufficient detail to facilitate replication. %
We introduce the SNN model by Diehl and Cook for unsupervised digit recognition~\cite{diehl2015unsupervised}, upon which we build our new system. The SNN contains biologically plausible mechanisms including Leaky-Integrate-and-Fire (LIF) neurons, Spike-Time Dependent Plasticity (STDP), lateral inhibition, and homeostasis, which enable unsupervised learning. %

The model consists of two layers, the input layer and the processing layer (\Cref{fig:motivation}). The input layer converts input images into spike trains using rate coding. The input layer contains one neuron per (resized) image pixel; i.e.~the number of neurons is $K_I=W \times H$, where $W$ and $H$ are the width and height of the input image, respectively. The input image is converted into Poisson-distributed spike trains by using each input pixel intensity as a firing rate (frequency of spikes) \cite{gerstner2014neuronal}.

The neurons in the input layer are fully connected (all-to-all connection) to all $K_P$ excitatory neurons in the processing layer. %
The processing layer also contains $K_P$ inhibitory neurons. The $k$-th inhibitory neuron receives input from the corresponding excitatory neuron $k$, and in turn, inhibits all excitatory neurons except the one that it receives the connection from. These inhibitory connections provide lateral inhibition enabling a model behavior similar to a winner-takes-all system.

The neuronal dynamics for all neurons are modeled using the LIF model, which is a common representation of biological spiking neuronal dynamics \cite{tieck2018controlling, BindsNET, renner2021backpropagation}. The differential equation describing the LIF dynamics (membrane voltage) is:
\begin{equation}
\tau \frac{dV}{dt} = (E_{rest} - V) + g_{e} (E_{exc} - V) + g_{i} (E_{inh} - V), 
\end{equation}
where $E_{rest}$ is the resting membrane potential, $E_{exc}$ and $E_{inh}$ are the excitatory and inhibitory synapses' equilibrium potentials with associated synaptic conductances $g_{e}$ and $g_{i}$.

The synaptic weights (representing synaptic strength) of the connections between the excitatory and inhibitory neurons are constant. The neuronal learning is done by adapting the synapses (connections) from the input layer to the excitatory neurons %
using STDP, which models the behavior of biological neurons where the magnitude and direction of change in synaptic strength (weight) is affected by the precise timing of presynaptic and postsynaptic spikes. Over a period of iterating over reference images in training, the synapses' weights increase the sensitivity of neurons to respond to different stimuli. In particular, all synaptic weights are modeled by the change in synaptic conductance. After a presynaptic input neuron sends a spike to the excitatory postsynaptic neuron, the synaptic connection is increased, otherwise, the conductance decays exponentially:
\begin{equation}
\tau_{ge} \frac{dg_{e}}{dt} = -g_{e},
\end{equation}
where the excitatory postsynaptic potential time constant is denoted by $\tau_{g_{e}}$. For synapses with an inhibitory presynaptic neuron, the same model is used to update conductance $g_{i}$, with inhibitory postsynaptic time constant $\tau_{gi}$.

Neuronal learning with STDP takes place by adapting the synaptic weights between the input layer and the excitatory neurons in the processing layer. The weights are modeled using presynaptic traces $\textit{x}_{pre}$, which are counters that record the number of presynaptic spikes. The change $\Delta w$ in synaptic weights after receiving a postsynaptic spike is modeled by:
\begin{equation}
    \Delta w = \eta (\textit{x}_{pre} - \textit{x}_{tar})(w_{max} - w)^\mu,
\end{equation}
where $\eta$ is the learning rate, $\textit{x}_{tar}$ is the value of the presynaptic trace when a postsynaptic spike is received, $w_{max}$ is the maximum weight, and $\mu$ is a ratio to represent the dependence of update on previous weight.

To ensure all output neurons respond to different patterns, and that the number of spikes fired by neurons for different labels remains approximately within a limited number, a bio-inspired mechanism, homeostasis, is deployed. Homeostasis is implemented by an adaptive neuronal threshold for excitatory neurons. The inner voltage threshold to raise a spike is increased by a constant $\theta$ after a neuron fires and decreases exponentially. For details about these mechanisms, we refer the interested reader to~\cite{diehl2015unsupervised} and our open-source code.

\section{Methodology: Neuronal Assignments}
\label{subsec:assignments}

Diehl and Cook~\cite{diehl2015unsupervised} considered the digit classification task, where each of the excitatory neurons unambiguously learned to represent one of ten digits (\Cref{subsec:hardassignments}).
This assumption 
does not hold for the visual place recognition task, where each place is considered as a separate class, resulting in a much larger number of classes\footnote{Note that we use the terms visual place recognition and place classification synonymously.}. 
Therefore, a significant number of neurons learn to represent more than a single place (i.e.~they fire to more than one place at training time). \Cref{subsec:weightedassignments} introduces our novel \emph{weighted} neuronal assignment strategy, taking a fine-grained account of the firing patterns at training time. Finally, \Cref{subsec:probability} defines a probability-based re-weighting scheme that increases the consistency between spike rates of the neuronal population.

\subsection{Standard Assignments}
\label{subsec:hardassignments}

The neuronal assignment scheme implemented by Diehl and Cook~\cite{diehl2015unsupervised} for digit recognition, which we refer to as standard assignments, is introduced here as follows as it serves as a baseline. The excitatory neurons were assigned to place labels based on the highest average response of the neurons to a place label in all training images $r\in R$. The place label $l$ is assigned to neuron $i$ if the sum of all spikes at training time was the highest for that label:

\begin{equation}
     N_{i} = \argmax_l S_{i,l}^R
\end{equation}
where $N_{i}$ is the assignment of neuron $i$ and $S_{i,l}^R$ is the total number of spikes of neuron $i$ when given images with label $l$ as input. 
We also define the set $M_l$ that contains all neurons that are assigned to label $l$, i.e.~$M_l=\{ i\ |\ N_i=l \}$.

At inference time, the predicted place label $l^*$ for a given query image $q$ is the place label $l$ that, summed over all neurons that were assigned to that label (i.e.~$m \in M_l$), had the highest spike rate:
\begin{equation}
     l^* = \argmax_l \sum_{m \in M_l} S_{m,l}^q.
\end{equation}

\subsection{Weighted Assignments}
\label{subsec:weightedassignments}

This section introduces our novel weighted neuronal assignment scheme. We considered each input frame as a new place, such that $R$ images of a traverse are considered as $R$ distinct places. 
With $K_P$ excitatory neurons to distinguish between $R$ places, experimental observation revealed that the majority of these output neurons can fire spikes in response to multiple similar places at different intervals in training. Of responsive neurons, there are three groups: 1) neurons that continuously only raise spikes to one particular place label, 2) neurons that raise spikes to two labels, and 3) neurons that fire spikes for many labels. 

However, Diehl and Cook's standard assignment scheme does not consider the responsiveness of neurons to multiple places. This led to our motivation for normalizing the neurons' spike rates according to the number of places that they have been assigned to at training time, in order to improve performance. Our proposed weighted assignments comprises of the following three strategies that work in conjunction. The strategies are applied one after the other and introduce just a single parameter $\gamma$ that is kept constant across all experiments.

\paragraph{Regularization by involvement}
The first normalization step lessens the contribution of neurons that have learned many place labels. Specifically, the spike rates of neurons that have learned more than $\gamma\%$ of place labels are regularized by the number of place labels that they have learned:

\begin{equation}
S_{i,l}^{Q'} = 
\begin{cases}
    S_{i,l}^Q & \text{if } \omega_i \leq \gamma\cdot R\\
    \frac{1}{\omega_i}S_{i,l}^Q,              & \text{otherwise}
\end{cases}
\end{equation}
where $S_{i,l}^{Q'}$ are the regularized spike rates and $\omega_i = \mathbf{card}\big(\left\{ i\ |\ N_i \in M_l \right\}\big)$ is the number of place labels represented by neuron $i$.

\paragraph{Normalization by response strength}
The next normalization step considers how strongly neuron $i$ has learned label $l$. 
This normalization step accounts for the strength of the responses; effectively measuring to what percentage label $l$ was learned compared to all labels $m$ that were learned\footnote{Note that most $S_{i,m}^R$ are zero.}:

\begin{equation}
S_{i,l}^{Q^*} = S_{i,l}^{Q'} \times \frac{S_{i,l}^R}{\sum\limits_m S_{i,m}^R}.
\end{equation}

\paragraph{Penalize relevant neurons that did not spike}
The last normalization step down-weights place labels $l$ when not all neurons $m\in M_l$ that have learned that place label, fired when seeing the query image. The normalization factor is the number of spikes (at training time) of the neurons that also fired at query time over the number of spikes of all neurons that have learned this label at training time:
 
\begin{equation}
\widehat{S_{i,l}}^{Q} = S_{i,l}^{Q^*} \times \frac{\sum\limits_m S_{i,m}^R \mathds{1}_{S_{i,m}^Q>0}}{\sum\limits_m S_{i,m}^R}.
\end{equation}

In our code repository, we show an ablation study that demonstrates that the combination of all three strategies leads to the best results, outperforming the use of single strategies or pairs of strategies.

\subsection{Probability-based Neuronal Assignments}
\label{subsec:probability}
Despite the addition of weighted assignments, the number of spikes of different query images varies across the dataset, complicating the assessment of whether a place match has been made. To account for the variation in spike numbers for different images, we converted the output spike numbers for every query image to a probability-based array. The probability-based spike numbers $\widetilde{S_{i,l}}^{Q}$ are achieved by min-max normalization scaled by the total number of spikes across all output neurons, as follows:
\begin{equation}
    \widetilde{S_{i,l}}^{Q} = \frac{1}{\sum\limits_k \widehat{S_{k,l}}^{Q}} \times \frac{ \widehat{S_{i,l}}^{Q} - \min\limits_k \widehat{S_{k,l}}^{Q} }{ \max\limits_k \widehat{S_{k,l}}^{Q} - \min\limits_k \widehat{S_{k,l}}^{Q}}.
\end{equation}

This method can be added as an additional step to both standard assignments and weighted assignments to increase consistency of spike rates for each neuron.

\section{Experimental Setup}

\begin{table*}[htbp]
\caption{Neuronal assignments: Mean recall at 100\% precision for the SNN model for $15 \times $100 place trials on Nordland, $4 \times $100 place trials on Oxford RobotCar, $6 \times $100 place trials on SPEDTest, and 100 place trials on Synthia and St Lucia each, using standard, weighted, probability-based, and weighted probability-based neuronal assignments.}
\renewcommand{\arraystretch}{1.2}

\label{Table_neuronal_assignments}
\centering
\begin{tabular}{c|cccccc}

Dataset & Standard & Weighted & Probability-based & Weighted probability-based & NetVLAD & SAD\\
\hline
Nordland & 12.8\% & 29.7\% & 38.2\% & 47.5\% & 15.3\% & 51.7\%\\

Oxford RobotCar & 2.5\% & 8.8\% & 4.8\% & 12.0\% & 11.0\% & 11.3\%\\ 

SPEDTest & 10.2\% & 19.8\% & 29.5\% & 18.7\% & 15.2\% & 7.7\%\\

Synthia & 1.0\% & 4.0\% & 22.0\% & 16.0\% & 15.0\% & 19.0\%\\ %

St Lucia & 12.0\% & 22.0\% & 13.0\% & 22.0\% & 6.0\% & 0.0\%
\end{tabular}
\vspace{-0.2cm}
\end{table*}

This section presents implementation details in \Cref{ER_I}, followed by the datasets and evaluation metrics used in \Cref{ER_D,ER_M} respectively.

\subsection{Implementation Details}
\label{ER_I}
We implemented the SNN model in Python3 using the Brian2 simulator~\cite{stimberg2019brian} (CPU version) and BindsNET~\cite{BindsNET} (GPU version). We base our implementation on \cite{diehl2015unsupervised}. 
All reference and query images were resized to $W\times H=28\times 28$ pixels. The images were also patch-normalized~\cite{milford2012seqslam} with patch sizes of $W_P\times H_P=7\times 7$ pixels to produce compact image representations.

The input images are converted from pixel values to Poisson spike trains using rate coding. This encoding method defines a firing rate for each input pixel based on its intensity. The pixel intensities are scaled to cap the frequencies of input spikes to a range of 0\,-\,63.75 Hz. The $K_I=784$ neurons in the first layer of the network each receive a spike train corresponding to the intensity of a single pixel in the input image. For the regularization term in the weighted assignments, we used a $\gamma$ value equal to 2\% of the number of reference places.

We trained the SNN model by presenting each input image for 350 ms and defined a resting period of 150 ms to allow neuronal dynamics to decay to their initial values. For a given image, the output of the model is the number of spikes raised by the $K_P=400$ excitatory neurons in the output layer. The model output is recorded after presenting each image. We trained the models on the reference dataset for 60 epochs.  

\subsection{Datasets}
\label{ER_D}
To evaluate the model performance, we used five widely established datasets that provide sequential images captured under different appearance conditions: Nordland~\cite{sunderhauf2013we}, Oxford RobotCar~\cite{RobotCar}, SPEDTest \cite{chen2018SPED}, Synthia \cite{ros2016synthia}, and St Lucia \cite{Glover2010StLucia}. We used the same model configuration to train all datasets and without any dataset-specific fine-tuning.

The Nordland dataset captures a train journey along a 728 km long route in Norway, with one traverse each in spring, summer, fall, and winter. As typically done in the literature~in~\cite{molloy2020intelligent,hausler2019multi}, all tunnels, stationary periods, and sections where the train travels below 15 km/h (based on GPS data) were removed. The frames have one-to-one correspondences across seasons, i.e.~frame $k$ in one season matches exactly frame $k$ in the other seasons. We used the spring and fall traverses as the reference dataset and the summer traverse as the query dataset. To increase independence between each place label, we sampled places every 8 seconds (roughly every 100 meters) from the first quarter of the data.

The Oxford RobotCar dataset contains over 100 traversals around Oxford, with variations in weather conditions, times of the day and seasons, and centimeter-accurate annotations. Following~\cite{molloy2020intelligent}, we selected the sun and rain traverses for reference, and the dusk traverse as the query dataset\footnote{Sun: 2015-08-12-15-04-18, rain: 2015-10-29-12-18-17, dusk: 2014-11-21-16-07-03}. We sampled places roughly every 10 meters.

The SPEDTest dataset consists of images obtained from surveillance cameras around the world under various weather, season, and illumination conditions with frame-to-frame correspondences \cite{chen2018SPED}. The Synthia dataset is a synthetic dataset in a city-like driving scenario, originally created for semantic scene understanding. As in \cite{zaffar2021vpr}, we used the SEQS-04 fog (reference) and night (query) traverses. We removed the sections where the vehicle is stopped and roughly sampled places at about 1 fps. 
The St Lucia dataset contains multiple traverses of a route through the St Lucia suburb, Brisbane. We used an early morning traverse (reference), and a late afternoon traverse (query)\footnote{Morning: 190809-0845, afternoon: 180809-1545}. We removed the sections where the vehicle is stationary and roughly sampled places every 20 meters. 

Since the performance of different methods varies significantly depending on the particular section of the dataset that is being used~\cite{schubert2021makes}, we performed comparisons on 15 non-overlapping sections on Nordland, 4 non-overlapping sections on Oxford RobotCar, 6 non-overlapping sections of SPEDTest, 1 section of Synthia, and 1 section of St Lucia.

\subsection{Evaluation Metrics}
\label{ER_M}
We use recall at 100\% precision (R@100P) and area under the precision-recall curve (AUC) to evaluate the performance of the SNN model, which is consistent with many works on VPR~\cite{cummins2008fab,labbe2013appearance,Lowry2015}. R@100P is an important metric when using VPR to propose loop closure candidates to Simultaneous Localization and Mapping (SLAM) systems, which ideally require maximal recall without false positives (i.e.~100\% precision). AUC is a convenient summary statistic. We impose a particularly tight ground truth tolerance, only considering exact matches as true matches, despite the relatively small gap between images.

\section{Results}
We first compare different neuronal assignment schemes on the loop closure detection task in \Cref{ER_NA}. This is followed by a comparison of our method with existing approaches for VPR, namely the widely used NetVLAD~\cite{Arandjelovic2018} and Sum-of-Absolute-Differences~\cite{milford2012seqslam} (\Cref{ER_SOTA}). Finally, we evaluate performance for a fixed network size when increasing the number of reference images in \Cref{ER_A}.

\begin{figure}[t!]
    \captionsetup[subfigure]{aboveskip=1.5pt,belowskip=0.5pt,labelformat=simple}
    \renewcommand*{\thesubfigure}{(\alph{subfigure})} %
    \centering
    \begin{subfigure}{0.49\columnwidth} %
        \centering
        \includegraphics[width=\textwidth]{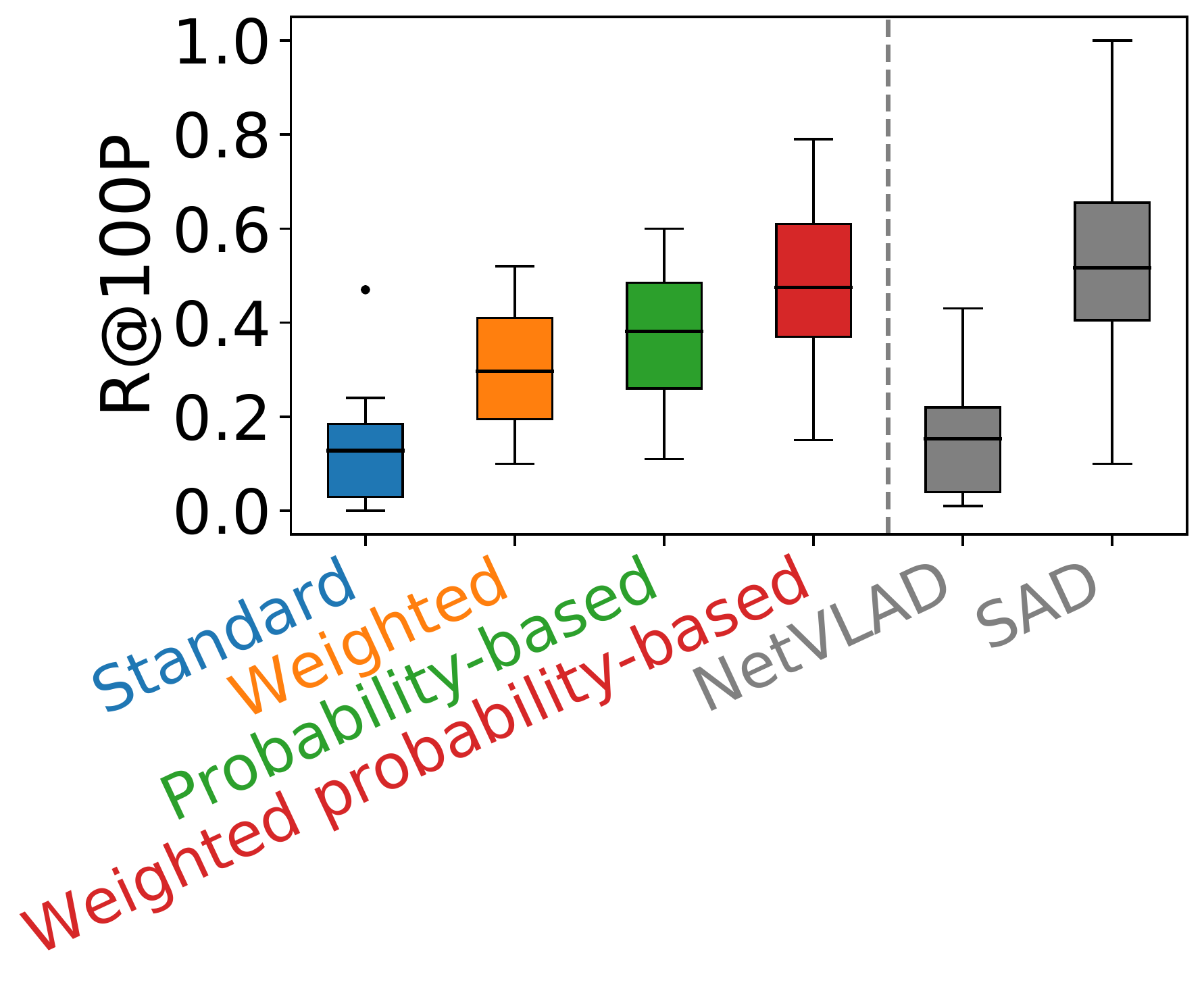}
        \caption{Nordland}
    \end{subfigure}
    \begin{subfigure}{0.49\columnwidth} %
        \centering
        \includegraphics[width=\textwidth]{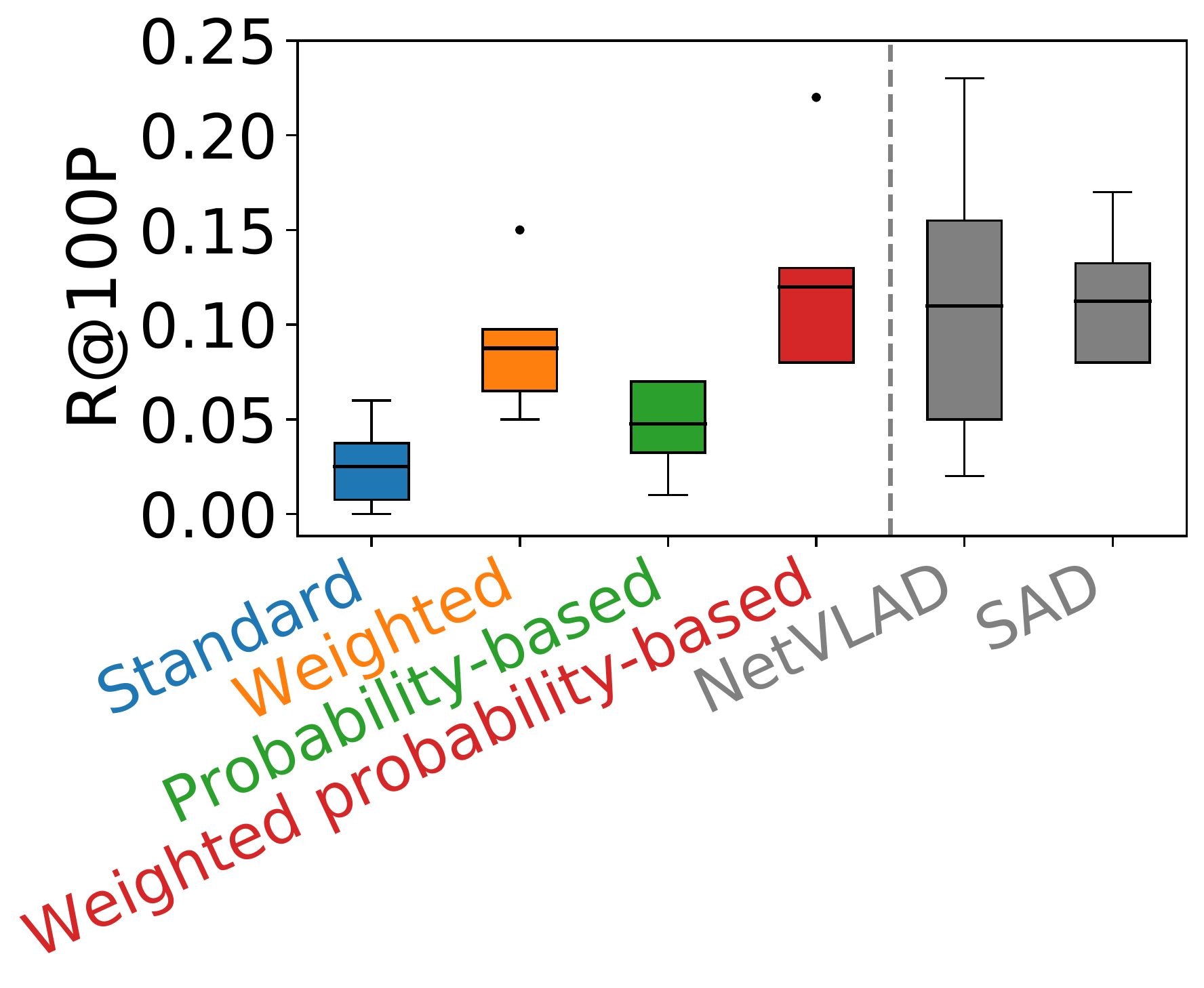}
        \caption{Oxford RobotCar}
    \end{subfigure}
    \caption{Comparison of recall at 100\% precision (R@100P) for various neuronal assignment schemes on Nordland and Oxford RobotCar datasets. The underlying data for the box plots stem from different sections of the dataset, where each section contains 100 places. The mean R@100P across all selected sections of datasets is indicated with a black line. For both datasets, our proposed neuronal assignment schemes lead to significant improvements over the baseline.}
    \label{fig:T_neuronal_assignments}
\end{figure}

\subsection{Loop Closure Detection} \label{ER_NA}

\Cref{Table_neuronal_assignments,fig:T_neuronal_assignments} present a comparison of the model performance using standard, weighted, probability-based, and weighted probability-based assignments on the Nordland, Oxford RobotCar, SPEDTest, Synthia, and St Lucia datasets. Compared to standard assignments, the weighted and probability-based assignments improve R@100P on average by 9.1\% and 13.8\% respectively (absolute increase). Our proposed combination of both of these methods, the weighted probability-based assignment, significantly improves the model performance on these datasets ($p<0.001$; paired $t$-test).

\begin{figure}
    \centering
    \captionsetup[subfigure]{aboveskip=1.5pt,belowskip=0.5pt,labelformat=simple}
    \renewcommand*{\thesubfigure}{(\alph{subfigure})}
    \begin{subfigure}{0.49\linewidth}
        \centering
        \includegraphics[width=\textwidth]{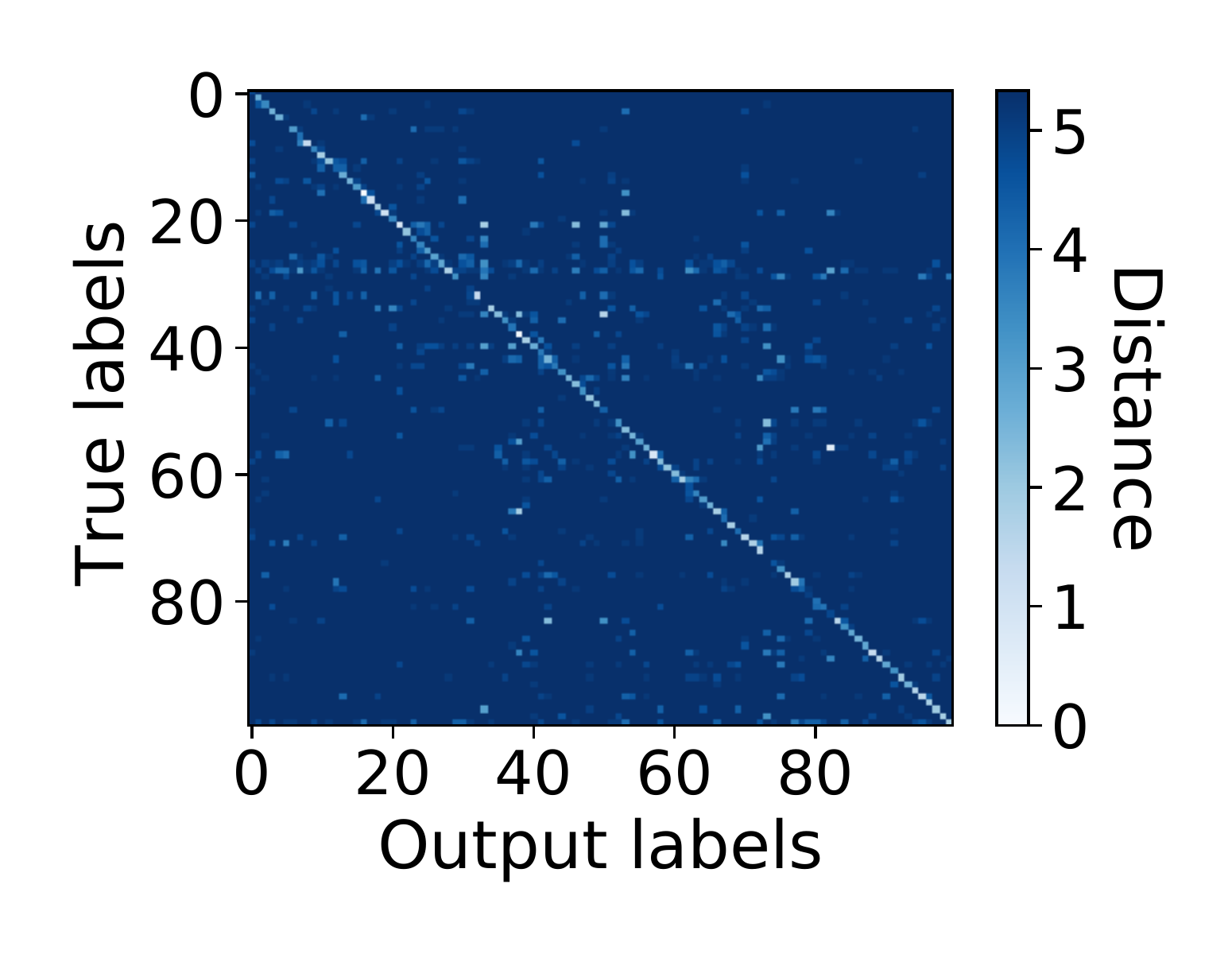}
        \caption{Standard}
    \end{subfigure}
    \begin{subfigure}{0.49\linewidth}
        \centering
        \includegraphics[width=\textwidth]{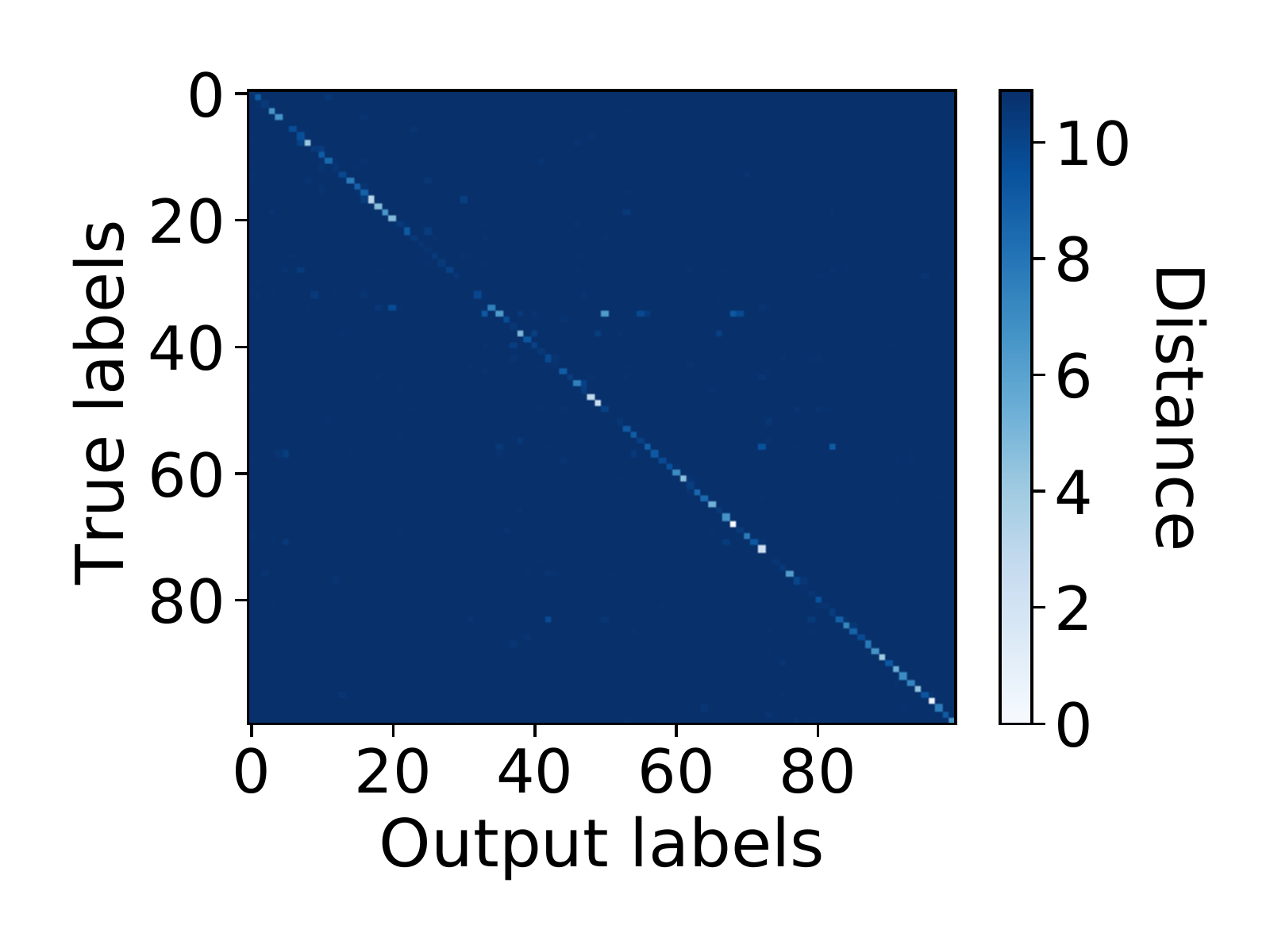}
        \caption{Weighted}
    \end{subfigure}

    \caption{Distance matrices of SNN model on a trial of Nordland with 100 places post-processed with standard and weighted neuronal assignments. The weighted assignments improve the model performance by upweighting neurons that have learned one label and downweighing neurons that have learned multiple labels. Compared to standard assignments, the weighted assignments decreases aliasing and maintains and/or improves P@100\%R and R@100\%P. }
    \label{fig:image_gap_DistMatrices}
\end{figure}

\Cref{fig:image_gap_DistMatrices} shows the distance matrices of the SNN model using 100 places from the Nordland dataset, obtained by subtracting the spike rates from the maximum observed spike rate (higher spike rates indicate higher similarities). Thus lower values indicate better matches; an ideal VPR system would have low values on the diagonal and high values elsewhere. In \Cref{fig:image_gap_DistMatrices}, the standard assignment achieved 78.0\% precision at 100\% recall (P@100R) and a R@100P of 2.1\%. The weighted assignments increased P@100R by 4.0\% and R@100P by 29.3\%.

\begin{figure}
    \begin{subfigure}{\columnwidth}
        \centering
        \includegraphics[width=0.65\textwidth]{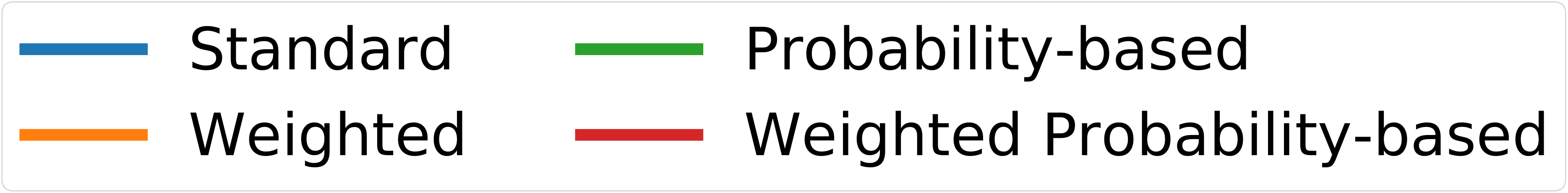}
        \hfill
    \end{subfigure}
    \captionsetup[subfigure]{aboveskip=1.5pt,belowskip=0.5pt,labelformat=simple}
    \renewcommand*{\thesubfigure}{(\alph{subfigure})} %
    \centering
    \begin{subfigure}{0.45\columnwidth}
        \centering
        \includegraphics[width=\textwidth]{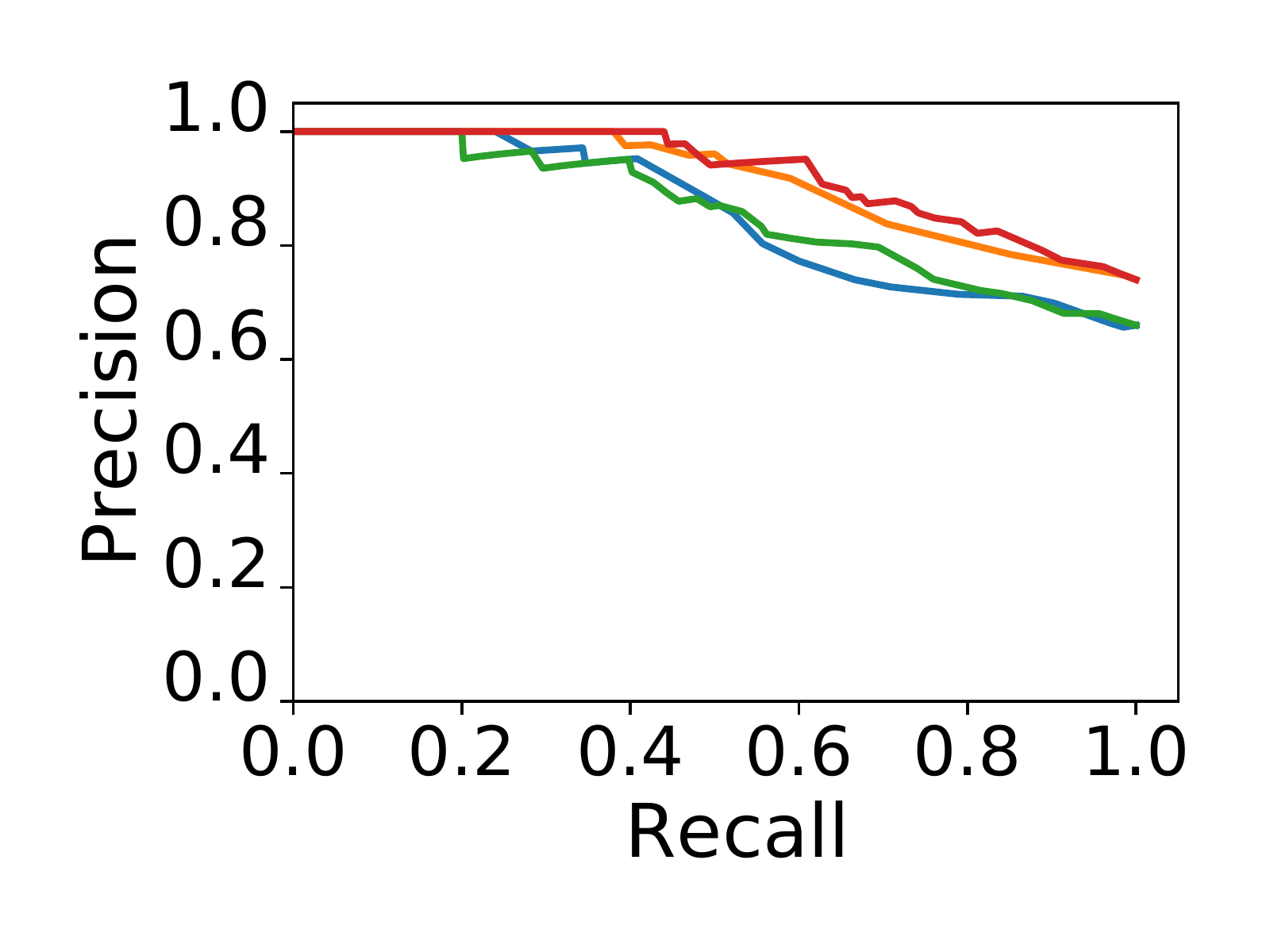}
        \caption{Nordland}
    \end{subfigure}
    \hfill
    \begin{subfigure}{0.45\columnwidth}
        \centering
        \includegraphics[width=\textwidth]{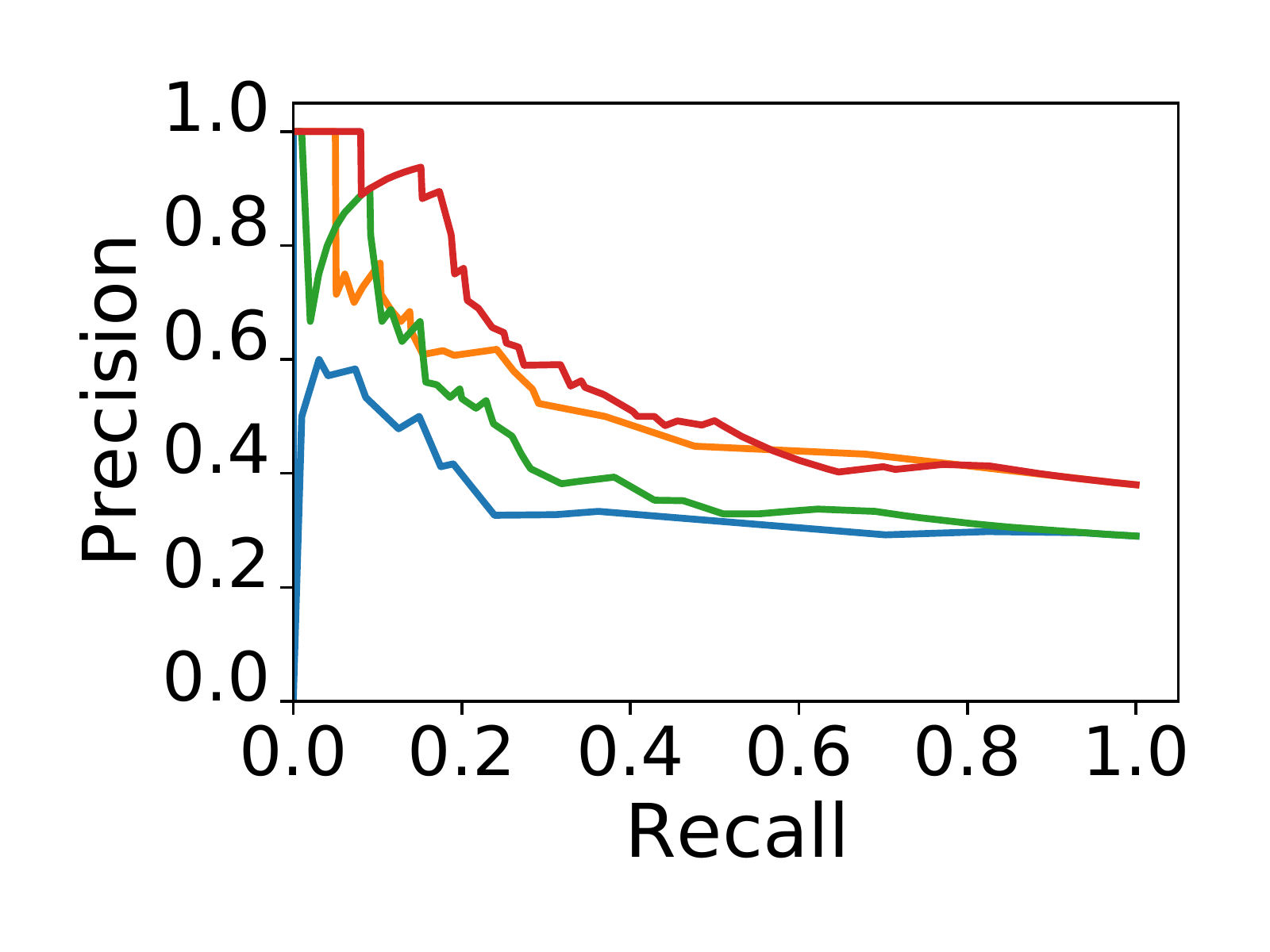}
        \caption{Oxford RobotCar}
    \end{subfigure}
    \caption{Comparison of SNN model post-processed with different assignments on Nordland and Oxford RobotCar datasets with 100 places: the weighted, and probability-based assignments can improve performance compared to the standard assignments. The combination of both approaches, weighted probability-based assignments outperforms the standard assignments.}
    \label{fig:PR_curves}
\end{figure}

\Cref{fig:PR_curves} presents the SNN model performance using standard, weighted, probability-based, and weighted probability-based assignments. Consistent with the results presented in \Cref{fig:T_neuronal_assignments,Table_neuronal_assignments}, the weighted and probability-based assignments increase recall at 100\% precision compared to the standard assignments. The weighted probability-based assignments outperforms standard assignments and maintains and/or improves performance compared to weighted and probability-based assignments.  

\begin{figure}[t]
\centering
\includegraphics[width=0.99\columnwidth]{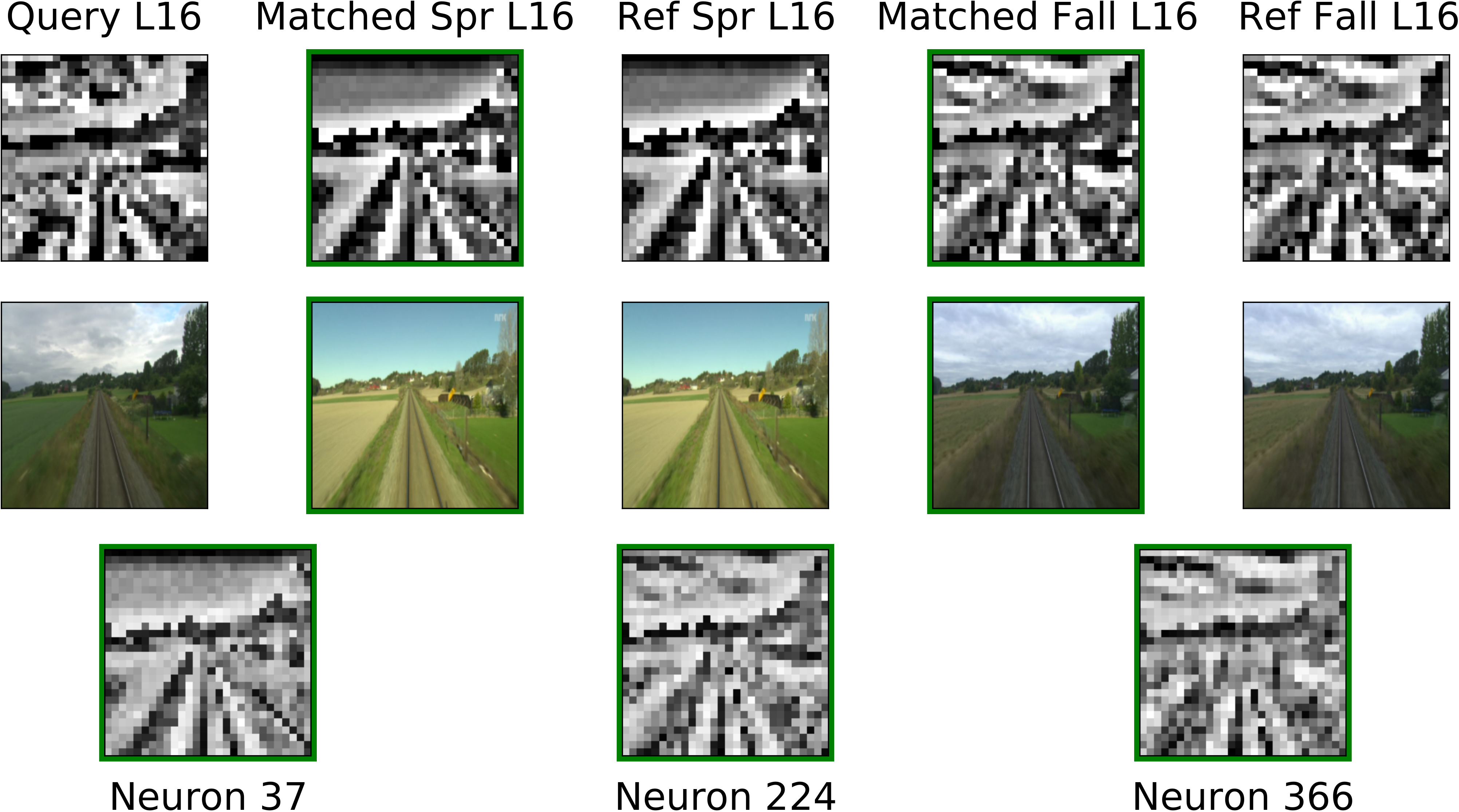}
\caption{Qualitative Results of our proposed SNN model with weighted assignments: An example of a query image from summer dataset correctly matched to its corresponding reference place from spring and fall traverses (indicated by green borders), in resized and patch normalized form, and original form. The last row shows the templates learned by the neurons assigned to this place.
}
\label{qualitative_results}
\end{figure}

\Cref{qualitative_results} shows a qualitative example; in particular, the query place, the place matched by the SNN model with weighted probability-based assignment, and the corresponding reference place for the given query image. The figure includes the templates learned by the neurons assigned to that place label.

\subsection{Comparison to State-of-the-art VPR Approaches} \label{ER_SOTA}
This section provides a comparison of our proposed SNN model with the standard VPR methods, namely the simplistic Sum of Absolute Differences (SAD)~\cite{milford2012seqslam} and a state-of-the-art approach, NetVLAD~\cite{Arandjelovic2018}. 

SAD measures the pixel-wise difference between the reference and query images. For a fair comparison, the input images of the SAD model were pre-processed with the same configuration as our SNN model (same image size and patch normalization; see \Cref{ER_I}). NetVLAD is known to generalize well across different datasets, having high robustness to both viewpoint and appearance changes. Due to the VGG backbone, NetVLAD requires input images to be at least 240x240 pixels, which is significantly higher than the input resolution to our SNN model and SAD.

\Cref{Table_neuronal_assignments,fig:T_neuronal_assignments} present a comparison of SNN model performance using standard, weighted, probability-based and weighted probability-based assignments as well as NetVLAD and SAD methods. For the Nordland dataset, the mean R@100P of the SNN model with weighted probability-based assignment (47.5\%) is comparable to the performance of SAD with an average R@100P of 51.7\%. We note that the performance of NetVLAD on the Nordland dataset is known to be relatively low due to the different training dataset~\cite{Garg2021,zaffar2021vpr}. For Oxford RobotCar dataset, the mean R@100P of the SNN model with weighted (8.8\%) and weighted probability-based (12.0\%) is comparable to that of NetVLAD (11.0\%) and SAD (11.3\%). The Oxford RobotCar dataset is challenging due to appearance change (sun and rain reference images and dusk queries) and occlusions.

For the SPEDTest dataset, our SNN model achieves a mean R@100P of 18.7\%, outperforming both NetVLAD (15.2\%) and SAD (7.7\%). 
For the Synthia dataset, our method achieves a R@100P of 16.0\%, which is similar to the performance of NetVLAD (15.0\% R@100P) and slightly worse than SAD (19.0\% R@100P). 
For the St Lucia dataset, our SNN method outperforms both NetVLAD (6.0\% R@100P) and SAD (0.0\% R@100P) with a R@100P of 22.0\%. 

We note that SAD performs well on the Nordland dataset as this dataset has minimum viewpoint change, which makes it well suited for SAD that calculates patch-wise differences to compute similarity (as highlighted in literature, some techniques are better suited than others for some datasets \cite{schubert2021makes}). 
In comparison, our learning-based approach can potentially generalize better to other datasets compared to SAD.

Getting an SNN-based model to not only mechanistically perform the VPR task, but also achieve broadly comparable performance to SAD and NetVLAD represents a substantial step forward. We note however that these existing techniques have been demonstrated on larger datasets, making further scalability (beyond the studies already presented here) one of the core foci of future work with the SNN-based approach.

\subsection{Ablation Study: Varying number of places} \label{ER_A}

In this section, we analyze the model performance (using our proposed weighted probability-based assignments) with a varying number of place labels and a lower number of epochs when increasing the number of places\footnotemark. This study covers the same section of the Nordland dataset which covers 400 place labels. Therefore, the model learning 25 place labels was repeated 16 times on consecutive sections ($16\times 25=400$), the model with 50 place labels was repeated 8 times, and so on. For the evaluation points where 400 is not divisible by the number of places, at least two sections were evaluated with varying offsets\footnotetext{24, 60, 60, 60, 30, 20, 20, 20 and 15 epochs for 25, 50, 100, 150, 200, 250, 300, 350, and 400 places respectively.}\footnote{250, 150, 100 and 50 offsets for 150, 250, 300 and 350 places respectively.}.

\begin{figure}[t]
\centering
\includegraphics[width=0.65\columnwidth]{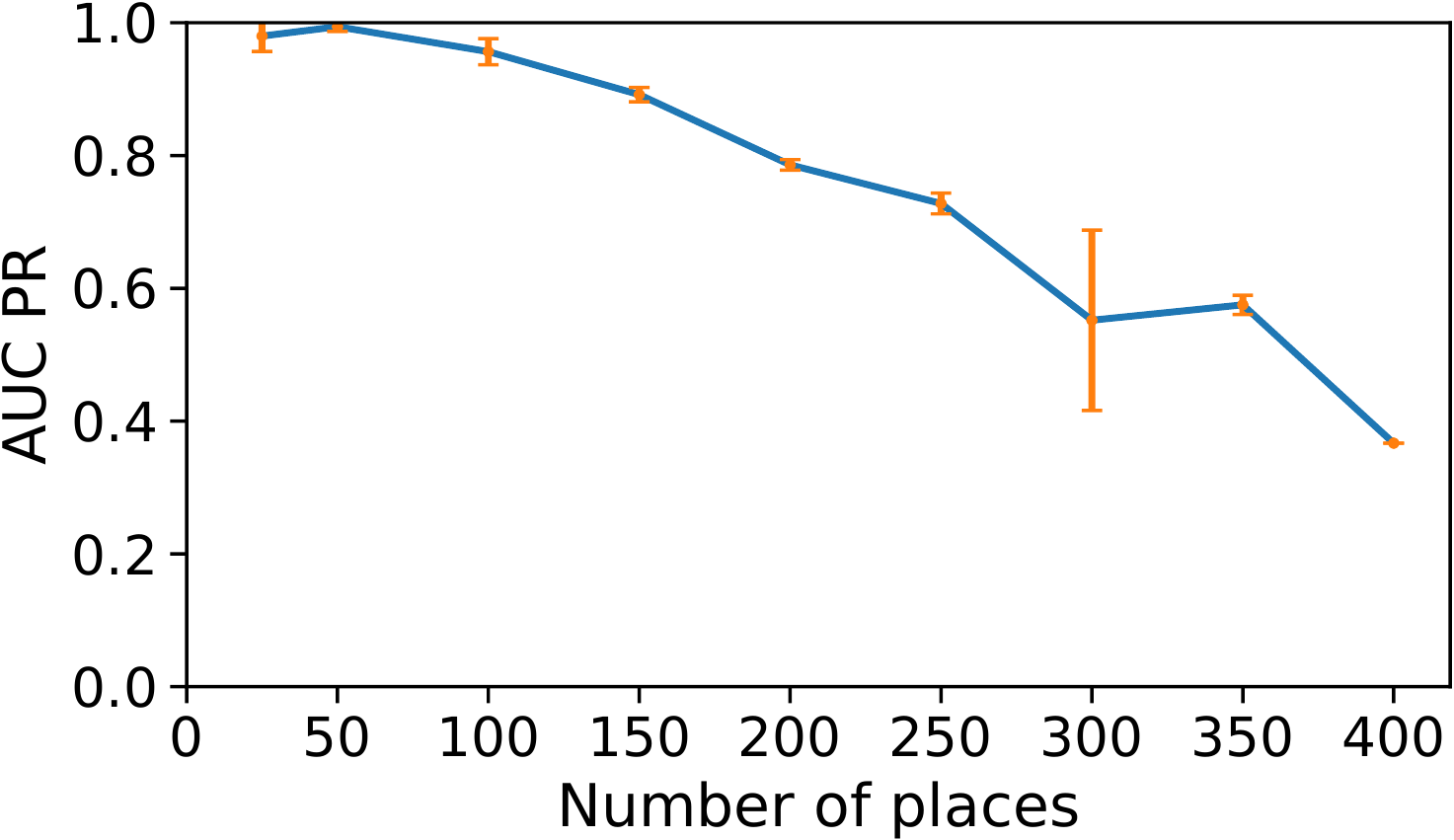}
\caption{SNN model performance with varying number of places (25, 50, 100, 150, 200, 250, 300, 350, 400). With a fixed network size, the model performance degrades gracefully with increasing number of places. 
}
\label{num_places}
\end{figure}

\Cref{num_places} shows that as the number of place labels increases, the model performance gracefully degrades when more than 100 places are learned. This is expected as the number of neurons assigned to a particular place label decreases with increasing number of places to encode, and the number of labels learnt by a neuron increases. Consequently, a single neuron outputs a higher number of spikes for \emph{different} place labels, resulting in worse performance.

As the number of places grows, a network with an increased number of neurons will be needed to maintain a certain desired level of performance. Characterising this relationship, we found that to achieve an average AUC of 92+/-2\%, the model approximately requires 100 output neurons for 50 places, 200 output neurons for 100 places, 450 output neurons for 150 places, and 800 output neurons for 200 places. However, the increase in the number of output neurons results in an approximately linear increase in inference time per query image (see our GitHub repository for figures). In \Cref{sec:conclusions}, we discuss our future work on approaches that can enable the network to maintain its performance for large-scale VPR.

An additional post-processing approach that can improve model performance is sequence matching as in SeqSLAM \cite{milford2012seqslam}.
The AUC of the model with 400 places is 39.9\% (\Cref{num_places}) using the weighted probability-based assignment (single-image based). By using SeqSLAM with a sequence length of 5 and 10 frames, the AUC of this model increases to 88.3\% and 98.4\% respectively, showing that our model is amenable to sequence processing (see our GitHub repository for a figure). We note that sequence matching can thus be used to implicitly encode a larger number of places. In future works, we would like to provide image sequences as input to our model instead of using a post-processing step.

\section{Discussions and Conclusions}
\label{sec:conclusions}
This research demonstrated that a two-layer SNN can be adapted for the place recognition task, through the introduction of a new weighted assignment scheme, and that it can achieve comparable performance to classical and state-of-the-art techniques on benchmark robotics datasets with significant degrees of appearance change.

Future work will pursue several promising directions. Firstly, we will investigate scaling the model size up to arbitrarily large environments, evaluating in particular how training and compute requirements scale and the resulting effect on performance. We will also investigate approaches to improve the robustness of our method to significant viewpoint changes. 

We are working with Intel to deploy this system on their new Loihi neuromorphic hardware. Such deployment has previously shown promise in terms of achieving high energy efficiencies with high throughput \emph{and} low latencies~\cite{frady2020neuromorphic}.
A number of recent works on SNNs \cite{frady2020neuromorphic, renner2021backpropagation} (which are loosely similar to ours and were deployed on Loihi) have reported inference times of less than 1 ms per query image, which is significantly lower than the inference time of our model on CPU (1 s) and GPU (0.2 s).  
The network upon which we base our system has been estimated to have less than 1 mW of power consumption on neuromorphic hardware \cite{diehl2015unsupervised}, which is significantly lower than the power consumption of our model on GPU (81 W). 

Loihi provides support to implement energy-efficient neuronal dynamics by supporting fine-grain parallelism, sparse event-based processing, and effective weight sharing \cite{davies2021advancing}. 
It will be feasible to deploy our SNN model on Intel's Kapoho Bay USB device which contains two Loihi chips and interfaces for sensor and event-based cameras \cite{davies2021advancing}.
Integrating the VPR capability into a full SNN-based SLAM system will assist in enabling the deployment of this system on low-power, low-cost robots with tight energy constraints.

Another avenue of future work includes optimization techniques for encoding mechanisms and structural synaptic plasticity mechanisms to improve learning speed, optimize computational resource allocation, and improve the encoding efficiency of place labels \cite{bogdan2018structural}. Finally, rather than converting intensity images to spike-trains, we are investigating the use of event cameras~\cite{Gallego2019} that directly provide spikes as output, providing the potential for a further reduction in energy consumption and localization latency. Previous works have demonstrated improvement in energy consumption and computational latency in robotic applications \cite{kreiser2018neuromorphic, kreiser2020error}.

\bibliographystyle{IEEEtran}
\bibliography{references}

\begin{thebibliography}{10}
\providecommand{\url}[1]{#1}
\csname url@rmstyle\endcsname
\providecommand{\newblock}{\relax}
\providecommand{\bibinfo}[2]{#2}
\providecommand\BIBentrySTDinterwordspacing{\spaceskip=0pt\relax}
\providecommand\BIBentryALTinterwordstretchfactor{4}
\providecommand\BIBentryALTinterwordspacing{\spaceskip=\fontdimen2\font plus
\BIBentryALTinterwordstretchfactor\fontdimen3\font minus
  \fontdimen4\font\relax}
\providecommand\BIBforeignlanguage[2]{{%
\expandafter\ifx\csname l@#1\endcsname\relax
\typeout{** WARNING: IEEEtran.bst: No hyphenation pattern has been}%
\typeout{** loaded for the language `#1'. Using the pattern for}%
\typeout{** the default language instead.}%
\else
\language=\csname l@#1\endcsname
\fi
#2}}

\bibitem{maass1997networks}
W.~Maass, ``Networks of spiking neurons: the third generation of neural network
  models,'' \emph{Neural Netw}, vol.~10, no.~9, pp. 1659--1671, 1997.

\bibitem{davies2021advancing}
M.~Davies \emph{et~al.}, ``Advancing neuromorphic computing with {Loihi}: A
  survey of results and outlook,'' \emph{Proc. IEEE}, vol. 109, no.~5, pp.
  911--934, 2021.

\bibitem{maass2015spike}
W.~Maass, ``To spike or not to spike: that is the question,'' \emph{Proc.
  IEEE}, vol. 103, no.~12, pp. 2219--2224, 2015.

\bibitem{WHITTINGTON2019235}
J.~C. Whittington and R.~Bogacz, ``Theories of error back-propagation in the
  brain,'' \emph{Trends Cogn. Sci.}, vol.~23, no.~3, pp. 235--250, 2019.

\bibitem{markram2012spike}
H.~Markram, W.~Gerstner, and P.~J. Sj{\"o}str{\"o}m, ``Spike-timing-dependent
  plasticity: a comprehensive overview,'' \emph{Front. Synaptic Neurosci.},
  vol.~4, no.~2, 2012.

\bibitem{tieck2017towards}
J.~C.~V. Tieck \emph{et~al.}, ``Towards grasping with spiking neural networks
  for anthropomorphic robot hands,'' in \emph{Int. Conf. Artificial Neural
  Netw.}, 2017, pp. 43--51.

\bibitem{tang2018gridbot}
G.~Tang and K.~P. Michmizos, ``Gridbot: an autonomous robot controlled by a
  spiking neural network mimicking the brain's navigational system,'' in
  \emph{Int. Conf. Neuromorphic Syst.}, 2018.

\bibitem{tieck2018controlling}
J.~C.~V. Tieck \emph{et~al.}, ``Controlling a robot arm for target reaching
  without planning using spiking neurons,'' in \emph{IEEE Int. Conf. Cogn.
  Informat. Cogn. Comput.}, 2018, pp. 111--116.

\bibitem{tang2019spiking}
G.~Tang, A.~Shah, and K.~P. Michmizos, ``Spiking neural network on neuromorphic
  hardware for energy-efficient unidimensional {SLAM},'' in \emph{IEEE/RSJ Int.
  Conf. Intell. Robot. Syst.}, 2019, pp. 4176--4181.

\bibitem{kreiser2018pose}
R.~Kreiser, A.~Renner, Y.~Sandamirskaya, and P.~Pienroj, ``Pose estimation and
  map formation with spiking neural networks: towards neuromorphic {SLAM},'' in
  \emph{IEEE/RSJ Int. Conf. Intell. Robot. Syst.}, 2018, pp. 2159--2166.

\bibitem{kreiser2020chip}
R.~Kreiser \emph{et~al.}, ``An on-chip spiking neural network for estimation of
  the head pose of the {iCub} robot,'' \emph{Front. Neurosci.}, vol.~14, no.
  551, 2020.

\bibitem{lele2021end}
A.~Lele, Y.~Fang, J.~Ting, and A.~Raychowdhury, ``An end-to-end spiking neural
  network platform for edge robotics: From event-cameras to central pattern
  generation,'' \emph{IEEE Trans. Cogn. Devel. Syst.}, 2021.

\bibitem{vitale2021event}
A.~Vitale, A.~Renner, C.~Nauer, D.~Scaramuzza, and Y.~Sandamirskaya,
  ``Event-driven vision and control for {UAVs} on a neuromorphic chip,'' in
  \emph{IEEE Int. Conf. Robot. Autom.}, 2021, pp. 103--109.

\bibitem{Lowry2015}
S.~Lowry, N.~S{\"{u}}nderhauf, P.~Newman, J.~J. Leonard, D.~Cox, P.~Corke, and
  M.~J. Milford, ``{Visual place recognition: A survey},'' \emph{IEEE Trans.
  Robot.}, vol.~32, no.~1, pp. 1--19, 2015.

\bibitem{Garg2021}
S.~Garg, T.~Fischer, and M.~Milford, ``{Where Is Your Place, Visual Place
  Recognition?}'' in \emph{Int. Joint Conf. Artif. Intell.}, 2021, pp.
  4416--4425.

\bibitem{Arandjelovic2018}
R.~Arandjelovic, P.~Gronat, A.~Torii, T.~Pajdla, and J.~Sivic, ``{NetVLAD: CNN}
  architecture for weakly supervised place recognition,'' \emph{IEEE Trans.
  Pattern Anal. Mach. Intell.}, vol.~40, no.~6, pp. 1437--1451, 2018.

\bibitem{milford2012seqslam}
M.~J. Milford and G.~F. Wyeth, ``{SeqSLAM}: Visual route-based navigation for
  sunny summer days and stormy winter nights,'' in \emph{IEEE Int. Conf. Robot.
  Autom.}, 2012, pp. 1643--1649.

\bibitem{sunderhauf2013we}
N.~S{\"u}nderhauf, P.~Neubert, and P.~Protzel, ``Are we there yet? {Challenging
  SeqSLAM} on a 3000 km journey across all four seasons,'' in \emph{IEEE Int.
  Conf. Robot. Autom. Worksh.}, 2013.

\bibitem{RobotCar}
W.~Maddern, G.~Pascoe, C.~Linegar, and P.~Newman, ``1 year, 1000 km: The
  {Oxford RobotCar} dataset,'' \emph{Int. J. Robot. Res.}, vol.~36, no.~1, pp.
  3--15, 2017.

\bibitem{chen2018SPED}
Z.~Chen, L.~Liu, I.~Sa, Z.~Ge, and M.~Chli, ``Learning context flexible
  attention model for long-term visual place recognition,'' \emph{IEEE Robot.
  Autom. Lett.}, vol.~3, no.~4, pp. 4015--4022, 2018.

\bibitem{ros2016synthia}
G.~Ros, L.~Sellart, J.~Materzynska, D.~Vazquez, and A.~M. Lopez, ``The
  {SYNTHIA} dataset: A large collection of synthetic images for semantic
  segmentation of urban scenes,'' in \emph{IEEE Conf. Comput. Vis. Pattern
  Recog.}, 2016, pp. 3234--3243.

\bibitem{Glover2010StLucia}
A.~Glover, W.~Maddern, M.~Milford, and G.~Wyeth, ``{FAB-MAP + RatSLAM:
  Appearance-based SLAM for Multiple Times of Day},'' in \emph{IEEE Int. Conf.
  Robot. Autom.}, 2010, pp. 3507--3512.

\bibitem{diehl2015unsupervised}
P.~U. Diehl and M.~Cook, ``Unsupervised learning of digit recognition using
  spike-timing-dependent plasticity,'' \emph{Front. Comput. Neurosci.}, vol.~9,
  no.~99, pp. 1--9, 2015.

\bibitem{zhu2020spatio}
L.~Zhu, M.~Mangan, and B.~Webb, ``Spatio-temporal memory for navigation in a
  mushroom body model,'' in \emph{Conf. Biomimetic Biohybrid Syst.}, 2020, pp.
  415--426.

\bibitem{cadena2016past}
C.~Cadena \emph{et~al.}, ``Past, present, and future of simultaneous
  localization and mapping: Toward the robust-perception age,'' \emph{IEEE
  Trans. Robot.}, vol.~32, no.~6, pp. 1309--1332, 2016.

\bibitem{milford2004ratslam}
M.~J. Milford, G.~F. Wyeth, and D.~Prasser, ``{RatSLAM}: a hippocampal model
  for simultaneous localization and mapping,'' in \emph{IEEE Int. Conf. Robot.
  Autom.}, 2004, pp. 403--408.

\bibitem{galluppi2012live}
F.~Galluppi \emph{et~al.}, ``Live demo: Spiking {RatSLAM: Rat} hippocampus
  cells in spiking neural hardware,'' in \emph{IEEE Conf. Biomed. Circuits
  Syst.}, 2012, p.~91.

\bibitem{steffen2020networks}
L.~Steffen \emph{et~al.}, ``Networks of place cells for representing 3d
  environments and path planning,'' in \emph{IEEE Int. Conf. Biomed. Robot.
  Biomechatronics}, 2020, pp. 1158--1165.

\bibitem{Gallego2019}
G.~Gallego \emph{et~al.}, ``{Event-based Vision: A Survey},'' \emph{IEEE Trans.
  Pattern Anal. Mach. Intell.}, vol.~44, no.~1, pp. 154--180, 2022.

\bibitem{milford2015place}
M.~Milford \emph{et~al.}, ``Place recognition with event-based cameras and a
  neural implementation of {SeqSLAM},'' \emph{arXiv:1505.04548}, 2015.

\bibitem{weikersdorfer2013simultaneous}
D.~Weikersdorfer, R.~Hoffmann, and J.~Conradt, ``Simultaneous localization and
  mapping for event-based vision systems,'' in \emph{Int. Conf. Comput. Vis.
  Syst.}, 2013, pp. 133--142.

\bibitem{vidal2018ultimate}
A.~R. Vidal, H.~Rebecq, T.~Horstschaefer, and D.~Scaramuzza, ``Ultimate slam?
  combining events, images, and imu for robust visual slam in hdr and
  high-speed scenarios,'' \emph{IEEE Robot. Autom. Lett.}, vol.~3, no.~2, pp.
  994--1001, 2018.

\bibitem{kreiser2018neuromorphic}
R.~Kreiser, M.~Cartiglia, J.~N. Martel, J.~Conradt, and Y.~Sandamirskaya, ``A
  neuromorphic approach to path integration: a head-direction spiking neural
  network with vision-driven reset,'' in \emph{IEEE Int. Symp. Circuits Syst.},
  2018.

\bibitem{kreiser2020error}
R.~Kreiser, G.~Waibel, N.~Armengol, A.~Renner, and Y.~Sandamirskaya, ``Error
  estimation and correction in a spiking neural network for map formation in
  neuromorphic hardware,'' in \emph{IEEE Int. Conf. Robot. Autom.}, 2020, pp.
  6134--6140.

\bibitem{renner2021backpropagation}
A.~Renner, F.~Sheldon, A.~Zlotnik, L.~Tao, and A.~Sornborger, ``The
  backpropagation algorithm implemented on spiking neuromorphic hardware,''
  \emph{arXiv:2106.07030}, 2021.

\bibitem{lee2020enabling}
C.~Lee, S.~S. Sarwar, P.~Panda, G.~Srinivasan, and K.~Roy, ``Enabling
  spike-based backpropagation for training deep neural network architectures,''
  \emph{Front. Neurosci.}, vol.~14, no. 119, pp. 1--22, 2020.

\bibitem{rueckauer2017conversion}
B.~Rueckauer, I.-A. Lungu, Y.~Hu, M.~Pfeiffer, and S.-C. Liu, ``Conversion of
  continuous-valued deep networks to efficient event-driven networks for image
  classification,'' \emph{Front. Neurosci.}, vol.~11, p. 682, 2017.

\bibitem{stockl2021optimized}
C.~St{\"o}ckl and W.~Maass, ``Optimized spiking neurons can classify images
  with high accuracy through temporal coding with two spikes,'' \emph{Nat.
  Mach. Intell.}, vol.~3, no.~3, pp. 230--238, 2021.

\bibitem{cummins2008fab}
M.~Cummins and P.~Newman, ``{FAB-MAP: Probabilistic} localization and mapping
  in the space of appearance,'' \emph{Int. J. Robot. Res.}, vol.~27, no.~6, pp.
  647--665, 2008.

\bibitem{jegou2010vlad}
H.~J{\'e}gou, M.~Douze, C.~Schmid, and P.~P{\'e}rez, ``Aggregating local
  descriptors into a compact image representation,'' in \emph{IEEE Conf.
  Comput. Vis. Pattern Recog.}, 2010, pp. 3304--3311.

\bibitem{hausler2021patch}
S.~Hausler, S.~Garg, M.~Xu, M.~Milford, and T.~Fischer, ``{Patch-NetVLAD}:
  Multi-scale fusion of locally-global descriptors for place recognition,'' in
  \emph{IEEE Conf. Comput. Vis. Pattern Recog.}, 2021, pp. 14\,141--14\,152.

\bibitem{yu2019spatial}
J.~Yu, C.~Zhu, J.~Zhang, Q.~Huang, and D.~Tao, ``Spatial pyramid-enhanced
  netvlad with weighted triplet loss for place recognition,'' \emph{IEEE Trans.
  Neural Netw. Learn. Syst.}, vol.~31, no.~2, pp. 661--674, 2019.

\bibitem{molloy2020intelligent}
T.~L. Molloy, T.~Fischer, M.~Milford, and G.~N. Nair, ``Intelligent reference
  curation for visual place recognition via {Bayesian} selective fusion,''
  \emph{IEEE Robot. Autom. Lett.}, vol.~6, no.~2, pp. 588--595, 2020.

\bibitem{frady2020neuromorphic}
E.~P. Frady \emph{et~al.}, ``Neuromorphic nearest neighbor search using
  {Intel's Pohoiki Springs},'' in \emph{Proc. Neuro-inspired Comput. Elements
  Worksh.}, 2020.

\bibitem{gerstner2014neuronal}
W.~Gerstner, W.~M. Kistler, R.~Naud, and L.~Paninski, \emph{Neuronal dynamics:
  From single neurons to networks and models of cognition}.\hskip 1em plus
  0.5em minus 0.4em\relax Cambridge University Press, 2014.

\bibitem{BindsNET}
H.~Hazan \emph{et~al.}, ``{BindsNET}: A machine learning-oriented spiking
  neural networks library in python,'' \emph{Front. Neuroinform.}, vol.~12,
  no.~89, 2018.

\bibitem{stimberg2019brian}
M.~Stimberg, R.~Brette, and D.~F. Goodman, ``Brian 2, an intuitive and
  efficient neural simulator,'' \emph{Elife}, vol.~8, p. e47314, 2019.

\bibitem{hausler2019multi}
S.~Hausler, A.~Jacobson, and M.~Milford, ``Multi-process fusion: Visual place
  recognition using multiple image processing methods,'' \emph{IEEE Robot.
  Autom. Lett.}, vol.~4, no.~2, pp. 1924--1931, 2019.

\bibitem{zaffar2021vpr}
M.~Zaffar \emph{et~al.}, ``{VPR-Bench}: An open-source visual place recognition
  evaluation framework with quantifiable viewpoint and appearance change,''
  \emph{Int. J. Comput. Vis.}, vol. 129, p. 2136–2174, 2021.

\bibitem{schubert2021makes}
S.~Schubert and P.~Neubert, ``What makes visual place recognition easy or
  hard?'' \emph{arXiv:2106.12671}, 2021.

\bibitem{labbe2013appearance}
M.~Labbe and F.~Michaud, ``Appearance-based loop closure detection for online
  large-scale and long-term operation,'' \emph{IEEE Trans. Robot.}, vol.~29,
  no.~3, pp. 734--745, 2013.

\bibitem{bogdan2018structural}
P.~A. Bogdan, A.~G. Rowley, O.~Rhodes, and S.~B. Furber, ``Structural
  plasticity on the spinnaker many-core neuromorphic system,'' \emph{Front.
  Neurosci.}, vol.~12, p. 434, 2018.

\end{thebibliography}

\end{document}